\pdfoutput=1
\documentclass[11pt,a4paper]{article}
\PassOptionsToPackage{dvipsnames}{xcolor}
\usepackage[hyperref]{emnlp2020}
\usepackage{times}
\usepackage{latexsym}
\usepackage{url}
\usepackage{booktabs}
\usepackage{tikz-dependency}
\usepackage{mathtools}
\usepackage{multirow}
\usepackage{array}
\usepackage{amsfonts}
\usepackage{amsmath}
\usepackage{inconsolata}
\usepackage{CJKutf8}

\usepackage{microtype}

\usepackage{gb4e}
\noautomath

\newtoggle{comment}
\toggletrue{comment}

\newcommand{\reftab}[1]{Table~\ref{#1}}
\newcommand{\reffig}[1]{Fig.~\ref{#1}}
\newcommand{\refsec}[1]{\S\ref{#1}}

\newcommand{\vecs}[1]{\ensuremath{\mathbf{#1}}}

\makeatletter
\newcommand*\dotp{\mathpalette\dotp@{.5}}
\newcommand*\dotp@[2]{\mathbin{\vcenter{\hbox{\scalebox{#2}{$\m@th#1\bullet$}}}}}
\makeatother

\newcommand{\chn}[1]{\begin{CJK*}{UTF8}{gbsn}{#1}\end{CJK*}}

\aclfinalcopy %
\def\aclpaperid{1957} %

\title{Semantic Role Labeling as Syntactic Dependency Parsing}

\author{
  Tianze Shi\thanks{~~Work done during an internship at Bloomberg L.P.}\\
  Cornell University \\
  {\tt tianze@cs.cornell.edu} \\\And
  Igor Malioutov\\
  Bloomberg L.P. \\
  {\tt imalioutov@bloomberg.net} \\\And
  Ozan {\.I}rsoy\\
  Bloomberg L.P. \\
  {\tt oirsoy@bloomberg.net} \\
}

\date{}

\begin{document}
\maketitle

\begin{abstract}

We reduce the task of (span-based) PropBank-style semantic role labeling (SRL)
to syntactic dependency parsing.
Our approach is motivated by our empirical analysis
that shows three common syntactic patterns account for
over $98\%$ of the SRL annotations for both English and Chinese data.
Based on this observation, we present a conversion scheme
that packs SRL annotations into dependency tree representations
through joint labels that permit highly accurate recovery back
to the original format.
This representation allows us to train statistical dependency parsers
to tackle SRL and achieve competitive performance
with the current state of the art.
Our findings show the promise of syntactic dependency trees
in encoding semantic role relations within their syntactic
domain of locality,
and point to potential further integration of syntactic methods
into semantic role labeling in the future.

\end{abstract}

\section{Introduction}
\label{sec:intro}

Semantic role labeling \citep[SRL;][]{palmer+10} analyzes texts
with respect to predicate argument structures
such as ``\emph{who} did \emph{what} to \emph{whom}, and \emph{how}, \emph{when} and \emph{where}''.
These generic surface semantic representations provide
richer linguistic analysis than syntactic parsing alone
and are useful in a wide range of downstream applications
including
question answering \citep{shen-lapata07,khashabi+18},
open-domain information extraction \citep{christensen+10},
clinical narrative understanding \cite{albright+13},
automatic summarization \citep{khan+15}
and machine translation \citep{liu-gildea10,xiong+12,bazrafshan-gildea13},
among others.

It is commonly acknowledged that syntax and semantics
are tightly coupled with each other \citep{levin-hovav05}.
In some forms of linguistic theories \citep{baker96,baker97},
semantic arguments are even hypothesized to be assigned under
consistent and specific syntactic configurations.
As a matter of practice, annotations of semantic roles
\citep[][\emph{inter alia}]{palmer+05} are typically
based on existing syntactic treebanks as an additional annotation layer.
Annotators are instructed \citep{babko-malaya+06,bonial+15}
to identify semantic arguments within the predicates'
domain of locality,\footnote{
The arguments can potentially be traces and null elements.
If a trace is selected as an argument,
it is automatically chained to its surface constituent
after syntactic movement.
}
respecting the strong connection between syntax and semantics.

\begin{figure}
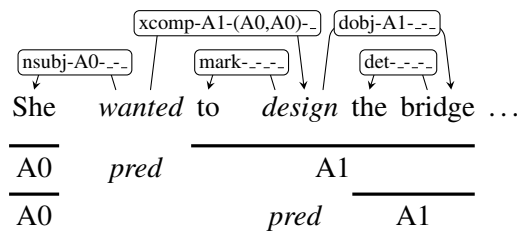

    \centering
    \begin{dependency}
    \begin{deptext}
    She \& \quad\emph{wanted}\quad \& to \& \quad\emph{design}\quad \& the \& bridge \& \ldots \\
    \end{deptext}
    \depedge[edge height=2ex]{2}{1}{nsubj-A0-\_-\_}
    \depedge[edge height=5ex, edge start x offset=5pt, edge end x offset=7pt]{2}{4}{xcomp-A1-(A0,A0)-\_}
    \depedge[edge height=2ex]{4}{3}{mark-\_-\_-\_}
    \depedge[edge height=5ex, edge start x offset=10pt, edge end x offset=5pt]{4}{6}{dobj-A1-\_-\_}
    \depedge[edge height=2ex]{6}{5}{det-\_-\_-\_}
    \node (a00l) [below left of = \wordref{1}{1}, xshift=1.4ex, yshift=1.2ex] {};
    \node (a00r) [below right of = \wordref{1}{1}, xshift=-1.4ex, yshift=1.2ex] {};
    \draw [-, very thick, black] (a00l) -- node[anchor=north] {A0} (a00r);

    \node (r0l) [below left of = \wordref{1}{2}, xshift=1.5ex, yshift=1.2ex] {};
    \node (r0r) [below right of = \wordref{1}{2}, yshift=1.2ex] {};
    \draw [-, very thick, white] (r0l) -- node[anchor=north, black] {\emph{pred}} (r0r);

    \node (a10l) [below left of = \wordref{1}{3}, xshift=2ex, yshift=1.2ex] {};
    \node (a10r) [below right of = \wordref{1}{6}, xshift=-0.5ex, yshift=1.2ex] {};
    \draw [-, very thick, black] (a10l) -- node[anchor=north] {A1} (a10r);

    \node (a01l) [below left of = \wordref{1}{1}, xshift=1.4ex, yshift=-2.5ex] {};
    \node (a01r) [below right of = \wordref{1}{1}, xshift=-1.4ex, yshift=-2.5ex] {};
    \draw [-, very thick, black] (a01l) -- node[anchor=north] {A0} (a01r);

    \node (a11l) [below left of = \wordref{1}{5}, xshift=2ex, yshift=-2.5ex] {};
    \node (a11r) [below right of = \wordref{1}{6}, xshift=-0.5ex, yshift=-2.5ex] {};
    \draw [-, very thick, black] (a11l) -- node[anchor=north] {A1} (a11r);

    \node (r1l) [below left of = \wordref{1}{4}, xshift=1.5ex, yshift=-2.5ex] {};
    \node (r1r) [below right of = \wordref{1}{4}, yshift=-2.5ex] {};
    \draw [-, very thick, white] (r1l) -- node[anchor=north, black] {\emph{pred}} (r1r);

    \end{dependency}
    \caption{An example sentence with SRL annotations (below) and our joint syntacto-semantic dependency relations (above; described in \refsec{sec:model}). The two representations can be converted from one to the other.
    A0 and A1 are short for SRL relations ARG0 and ARG1.}
    \label{fig:example}
\end{figure}

Empirically, syntax has indeed been shown to be helpful
to SRL in a variety of ways.
Earlier SRL systems have successfully incorporated
syntactic parse trees as features and pruning signals \citep{punyakanok+08}.
Recently, neural models with shared representations
trained to predict both syntactic trees and predicate-argument structures
in a multi-task learning setting achieve superior performance
to syntax-agnostic models \citep{strubell+18,swayamdipta+18a},
reinforcing the utility of syntax in SRL.

However, researchers are yet to fully leverage
all the theoretical linguistic assumptions
and the dataset annotation conventions
surrounding the tight connections between syntax and SRL.
To do so, ideally, one must perform
deep syntactic processing
to capture long-distance dependencies and argument sharing.
One solution is to introduce traces into phrase-structure trees,
which, unfortunately, is beyond the scope of
most statistical constituency parsers
partially due to their associated increased complexity
\citep{kummerfeld-klein17}.
Another solution is to use richer grammar formalisms
with feature structures
such as combinatory categorial grammar \citep[CCG;][]{steedman00}
and tree adjoining grammar \citep[TAG;][]{joshi+75}
that directly build syntactic relations
within the predicates' \emph{extended} domain of locality.
It is then possible to restrict the semantic argument candidates
to only those ``local'' dependencies \citep{gildea-hockenmaier03,liu09,liu-sarkar09,konstas+14,lewis+15}.
However, such treebank data are harder to obtain,
and their parsing algorithms tend to be less efficient
than parsing probabilistic context-free grammars \citep{kallmeyer10}.

On the other hand, syntactic dependency trees
directly encode bilexical governor-dependent relations
among the surface tokens,
which implicitly extend the domain of locality \citep{schneider08}.
Dependency parsing \citep{kubler+08} is empirically
attractive for its simplicity, data availability,
efficient and accurate parsing algorithms,
and its tight connection to semantic analysis \citep{reddy+17}.
Despite ample research community interest
in joint models for dependency parsing and SRL \citep{surdeanu+08,hajic+09,henderson+13},
a precise characterization of the mapping between semantic arguments
and syntactic configurations has been lacking.
In this paper, we provide a detailed empirical account of
PropBank-style SRL annotations on both English and Chinese data.
We show that a vast majority (over $98\%$) of the semantic relations
are characterized by one of three basic dependency-based syntactic configurations:
the semantic predicate 1) directly dominates, 2) is directly dominated by,
or 3) shares a common syntactic governor with the semantic argument.
The latter two cases are mostly represented by
syntactic constructions including relativization, control, raising, and coordination.

Based on our observations,
we design a back-and-forth conversion algorithm
that embeds SRL relations into dependency trees.
The SRL relations are appended to the syntactic labels
to form joint labels,
while the syntactic governor for each token remains unaltered.
The algorithms reach over $99\%$ F1 score on English
and over $97\%$ on Chinese data in oracle back-and-forth conversion experiments.
Further, we train statistical dependency parsing models
that simultaneously predict SRL and dependency relations
through these joint labels.
Experiments show that our fused syntacto-semantic models
achieve competitive performance with the state of the art.

Our findings show the promise of dependency trees
in encoding PropBank-style semantic role relations:
they have great potential in reducing the task of SRL
to dependency parsing with an expanded label space.
Such a task reduction facilitates future research
into finding an empirically adequate granularity
for representing SRL relations.
It also opens up future possibilities for further integration of
syntactic methods into SRL as well as
adaptations of extensively-studied dependency parsing techniques to SRL,
including linear-time decoding,
efficiency-performance tradeoffs,
multilingual knowledge transfer, and more.
We hope our work can inspire future research into
syntactic treatment of other shallow semantic representations
such as FrameNet-style SRL \citep{baker+98,fillmore+03}.
Our code is available at \url{https://www.github.com/bloomberg/emnlp20\_depsrl}.

\paragraph{Contribution}
Our work (1)
provides a detailed empirical analysis of
the syntactic structures of semantic roles,
(2) characterizes the tight connections between syntax and SRL
with three repeating structural configurations,
(3) proposes a back-and-forth conversion method
that supports a fully-syntactic approach to SRL,
and (4) shows through experiments that dependency parsers
can reach competitive performance
with the state of the art on span-based SRL.
Additionally,
(5) all our analysis, methods and results
apply to two languages from distinctive language families,
English and Chinese.

\section{Syntactic Structures of Semantic Roles}
\label{sec:pilot}

It has been widely assumed in linguistic theories
that the semantic representations of arguments
are closely related to their syntactic positions
with respect to the predicates
\citep{gruber65,jackendoff72,jackendoff92,fillmore76,baker85,levin93}.\footnote{
This is often termed \emph{linking theory} in linguistics (See \citet{levin-hovav05} for a survey).
}
This notion is articulated as linguistic hypotheses underlying
many syntactic theories:
\begin{exe}
\ex
\label{ex:uah}
\emph{Universal Alignment Hypothesis:}
There exist principles of Universal Grammar
which predict the initial [grammatical] relation borne
by each nominal in a given clause from the meaning of the clause.
\citep[p.~97]{perlmutter-postal84}
\ex
\label{ex:utah}
\emph{The Uniformity of Theta Assignment Hypothesis:}
Identical thematic relationships between items are
represented by identical structural relationships between
those items at the level of D[eep]-structure.
\citep[p.~57]{baker85}
\end{exe}
For theories that posit one-to-one correspondence
between semantic roles and syntactic structures \citep{baker96,baker97},
SRL can be treated purely as a syntactic task.
However, doing so would require deep structural analysis \citep{bowers10}
that hypothesizes more functional categories than what current syntactic annotations cover.

Nonetheless, the Proposition Bank \citep[PropBank;][]{kingsbury-palmer02,palmer+05}
annotations do capture the domain of locality that is implicitly
assumed by these linguistic theories.
PropBank defines the domain of locality for verbal predicates
to be indicated by ``clausal boundary markers''
and the annotators are instructed to limit their semantic role annotations
to ``the sisters of the verb relation (for example, the direct object)
and the sisters of the verb phrase (for example, the subject)''
\citep[p.~746]{bonial+17}.
In cases of syntactically-displaced arguments,
the annotators are asked to pick the empty elements
that are within the domain of locality,
and then syntactic coindexation chains
are used to reconstruct the surface semantic role relations.
Recognizing displaced arguments is crucial to SRL,
so taking full advantage of locality constraints
would also require modeling empty elements and movement,
for which current NLP systems still lack
accurate, efficient, and high-coverage solutions \citep{gabbard+06,kummerfeld-klein17}.

From an empirical perspective,
most syntactic realizations for semantic arguments
follow certain common patterns even when they are displaced.
Indeed, this is partially
why syntax-based features and candidate pruning heuristics
have been successful in SRL \cite{gildea-palmer02,gildea-jurafsky02,sun+08}.
Full parsing might not be necessary to account
for the majority of cases in the annotations.
Thus, knowing the empirical distributions
of the arguments' syntactic positions
would be highly useful for deciding
how detailed the syntactic analysis
needs to be for the purpose of SRL.
In this section,
we provide such a
characterization.

Our analysis is based on dependency syntax
and complements prior constituent-based characterizations \citep{palmer+05}.
One advantage of syntactic dependencies over phrase-structure trees
for the purposes of this paper is that
the dependents are often more directly connected to the syntactic governors
without intervening intermediate constituents.
For example, when a verb has multiple adjunct modifiers,
each would create an additional intermediate VP constituent
in the argument structure,
leading to further separation
between the verb and the external argument (subject).
In contrast,
in a dependency representation,
the subject is always directly dominated by the verbal predicate.

\subsection{Material}

\begin{table*}[t]
\centering
\small
\begin{tabular}{c>{\centering\arraybackslash}m{80pt}|m{220pt}|rr}
\toprule
&
\multicolumn{1}{c|}{\multirow{2}{*}{Pattern}}
&
\multicolumn{1}{c|}{\multirow{2}{*}{Example}}
&
\multicolumn{2}{c}{Percentage}
\\
&
&
&
\multicolumn{1}{c}{English}
&
\multicolumn{1}{c}{Chinese}
\\
\midrule
\multirow{1}{*}{(D)}
&
\begin{dependency}
\begin{deptext}
pred \& arg \\
\end{deptext}
\depedge[hide label,edge height=1.8ex]{1}{2}{}
\end{dependency}
&
\begin{dependency}
\begin{deptext}
\textcolor{MidnightBlue}{\emph{She}} \& \textcolor{BrickRed}{\emph{designed}} \& the \& bridge \& \ldots \\
\end{deptext}
\depedge[edge height=1.8ex]{2}{1}{nsubj}
\end{dependency}
&
$87.5\%$
&
$82.7\%$
\\
(C)
&
\begin{dependency}
\begin{deptext}
arg \& \hspace{20pt} \& pred \\
\end{deptext}
\depedge[hide label,edge height=1.8ex]{2}{1}{}
\depedge[hide label,edge height=1.8ex]{2}{3}{}
\end{dependency}
&
\begin{dependency}
\begin{deptext}
\textcolor{MidnightBlue}{\emph{She}} \& wanted \& to \& \textcolor{BrickRed}{\emph{design}} \& the \& bridge \& \ldots \\
\end{deptext}
\depedge[edge height=1.8ex]{2}{1}{nsubj}
\depedge[edge height=1.8ex]{2}{4}{xcomp}
\end{dependency}
&
$6.1\%$
&
$10.4\%$
\\
(R)
&
\begin{dependency}
\begin{deptext}
arg \& pred \\
\end{deptext}
\depedge[hide label,edge height=1.8ex]{1}{2}{}
\end{dependency}
&
\begin{dependency}
\begin{deptext}
The \& \textcolor{MidnightBlue}{\emph{bridge}}, \& which \& is \& \textcolor{BrickRed}{\emph{designed}} \& by \& her, \& \ldots \\
\end{deptext}
\depedge[edge height=1.8ex]{2}{5}{rcmod}
\end{dependency}
&
$4.7\%$
&
$5.7\%$
\\
&
\begin{dependency}
\begin{deptext}
arg \& \hspace{10pt} \& \hspace{10pt} \& pred \\
\end{deptext}
\depedge[hide label,edge height=1.8ex]{2}{1}{}

\depedge[hide label,edge height=1.8ex]{2}{3}{}
\depedge[hide label,edge height=1.8ex]{3}{4}{}
\end{dependency}
&
\begin{dependency}
\begin{deptext}
\textcolor{MidnightBlue}{\emph{She}} \& wanted \& to \& design \& and \& \textcolor{BrickRed}{\emph{build}} \& the \& bridge \& \ldots \\
\end{deptext}
\depedge[edge height=2ex]{2}{1}{nsubj}
\depedge[edge height=2ex]{2}{4}{xcomp}
\depedge[edge height=2ex]{4}{6}{conj}
\end{dependency}
&
$1.1\%$
&
$1.0\%$
\\
\midrule
\multicolumn{3}{c|}{Others}
&
$0.5\%$
&
$0.2\%$
\\
\bottomrule
\end{tabular}
\caption{
    The most common structural relations in the training data
    between the \textcolor{BrickRed}{\emph{predicates}} (pred) and the \textcolor{MidnightBlue}{\emph{arguments}} (arg).
    Appendix \refsec{app:eng} and \refsec{app:chinese} include
    more examples as well as Chinese data.
}
\label{tab:patterns}
\end{table*}

We use the training splits of the CoNLL 2012 shared task data
\citep{pradhan+12} on both English and Chinese;
sentences are originally from OntoNotes 5.0 \citep{hovy+06}.
The SRL annotations are based on English and Chinese PropBank
\citep{kingsbury-palmer02,palmer+05,xue-palmer03,xue08},
which are extensively used in SRL research.
We choose not to use the SRL-targeted CoNLL 2005
shared task \citep{carreras-marquez05} data
since earlier versions of PropBank \citep{babko-malaya05}
contain many resolvable mismatches between syntactic and semantic annotations
\citep{babko-malaya+06}.
Updated annotation guidelines \cite{bonial+15}
have fixed most of the identified issues.
We convert the Penn TreeBank \citep[PTB;][]{marcus+93}
and the Penn Chinese TreeBank \citep[CTB;][]{xue+05}
phrase-structure trees into Stanford Dependencies
\citep[SD;][]{demarneffe+06}
for English \citep{demarneffe-manning08,silveira+14}
and for Chinese \citep{chang+09} respectively.\footnote{
During conversion, we set copular verbs to be heads,
since PropBank marks some copular verbs as predicates.
}
SD is semantically-friendly
as noted by \citet[p.~2371]{schuster-manning16},
``Since its first version, SD representation has had
the status of being both a syntactic and a shallow semantic representation'',
thus it is suited for the development of our joint modeling
of syntactic and semantic structures.
Indeed, Universal Dependencies \citep[UD;][]{nivre+16},
which builds upon SD,
has been compared with and aligned to
meaning representations including UCCA \citep{hershcovich+19} and AMR \citep{szubert+18}.\footnote{
Our choice of SD instead of UD
is motivated by the flexibility in conversion to
set copular verbs as syntactic heads.
}

\subsection{Observations}
\label{sec:pilot-findings}

We categorize the syntactic configurations
between predicates and arguments
and present the results in \reftab{tab:patterns}.
For both English and Chinese, the vast majority,
more than $98\%$, of the predicate-argument relations
fall into one of three major categories:
the semantic argument is a syntactic child,
sibling, or parent of the semantic predicate.
Next, we give a brief account of our linguistic observations
on the English data associated with each category.
See Appendix \refsec{app:eng} and \refsec{app:chinese}
for more examples from both English and Chinese.

\paragraph{pred $\rightarrow$ arg (D)}
The predicate directly (D) dominates the semantic argument
in the syntactic tree.
Not surprisingly, this straightforward type of relation
is the most prevalent in the PropBank data,
accounting for more than $87\%$ ($82\%$) of
all English (Chinese) predicate-argument relations.

\paragraph{arg $\leftarrow$ $\rightarrow$ pred (C)}
The predicate and the argument share a common (C) syntactic parent.
There are two major types of constructions
resulting in this kind of configuration:
1) the common parent is a control or raising predicate,
creating an open clausal complement (xcomp) relation
and 2) there is a coordination structure
between the predicate and the common parent and
both predicates share a same argument in the semantic structure.
Both cases are so common that they are converted
to direct dependencies in the enhanced Stanford Dependencies \citep{schuster-manning16}.

\paragraph{arg $\rightarrow$ pred (R)}
The dominance relation between the predicate and the argument
is reversed (R).
This type of relations is frequently realized
through relative clauses (rcmod)
and verb participles (e.g., \emph{broken glass}).

\paragraph{Other constructions}
Many other constructions can be analyzed as combinations
of the previously mentioned patterns.\footnote{
Combinations of (D), (C), and (R) can theoretically account for
all possible predicate-argument configurations.
However, for a lossless back-and-forth conversion
with our proposed joint labels (\refsec{sec:model}),
there are constraints on the argument structures of all the intermediate predicates
along the shortest dependency path between the predicate and the argument.
See \reftab{tab:oracle} for an estimation of
how many semantic relations may be decomposed as
combinations of the three common structural patterns
empirically given our conversion method.
}
For example, a combination of (C)+(C)
through control and coordination
would derive the structural configuration of
the fourth most frequent case in \reftab{tab:patterns}.

\section{Reducing SRL to Dependency Parsing}
\label{sec:model}

\subsection{Joint Labels}
\label{sec:joint}

Building on the insights obtained from our analysis,
we design a joint label space to encode
both syntactic and SRL relations.
The joint labels have four components:
one syntactic relation
and three semantic labels, each corresponding
to one of the three most common structural patterns
in \reftab{tab:patterns}.

Formally, for a length-$n$ input sentence $w=w_1,\ldots,w_n$,
we denote the head of token $w_i$ in the syntactic dependency tree $t$
to be $w_{h_i}$, or $h_i$ for short.
The dependency tree also specifies a dependency relation labeled
$r_i$ between each ($h_i$, $w_i$) pair.
To encode both syntactic and SRL information,
we define a dependency tree $t'$,
keeping all the $h_i$'s same as in $t$,
but we modify relation $r_i$ to be
$r_i'\coloneqq r_i^{\textsc{SYN}}$-$r_i^{\text{(D)}}$-$r_i^{\text{(C)}}$-$r_i^{\text{(R)}}$,
a concatenation of four labels:
$r_i^{\textsc{SYN}}=r_i$ is the syntactic relation;
$r_i^{\text{(D)}}$ describes the SRL relation directly
between the predicate $h_i$ and the argument headed by $w_i$;
$r_i^{\text{(R)}}$ specifies the reverse situation
where $w_i$ is the predicate and $h_i$ the head of the argument;
$r_i^{\text{(C)}}$ encodes the parent-sharing pattern
connecting the two predicates and is in the form of
a tuple $(a, b)$, corresponding to the case
where the SRL argument with label $a$ for predicate $h_i$
is an SRL argument labeled $b$ with respect to predicate $w_i$.\footnote{
This assumes that
the argument must also be an argument to the predicate $h_i$.
In cases where there exists no such relation,
we insert a dummy relation $\emptyset$
that gets removed during post-processing
between $h_i$ and the argument,
and the (C) label between $h_i$ and $w_i$
then becomes $(\emptyset, b)$.
}
If there exist no such semantic relations,
the component labels can be left unspecified, denoted as ``\_''.

In the example of \reffig{fig:example},
the joint label between \emph{wanted} and \emph{design}
is xcomp-ARG1-(ARG0,ARG0)-\_.
We can break the joint label into four parts:
``xcomp'' describes the syntactic relation between the two tokens;
``ARG1'' indicates that the subtree \emph{to design the bridge}
is an argument labeled ``ARG1'' for predicate \emph{wanted};
(ARG0,ARG0) establishes the argument sharing strategy that
ARG0 \emph{she} of \emph{wanted} is an ARG0 for the predicate \emph{design};
finally, ``\_'' indicates there is
no argument headed by \emph{wanted} for the predicate \emph{design}.

\subsection{Back-and-Forth Conversion}
\label{sec:conversion}

The joint labels encode both syntactic and semantic relations,
and it is straightforward to convert/recover
the separate dependency and SRL annotations
to/from the joint representations.

In the forward conversion (separate $\rightarrow$ joint),
we first extract the syntactic heads of all SRL arguments.
Then we enumerate all predicate-argument pairs,
and for each pair falling into one of the three most common patterns
as listed in \reftab{tab:patterns},
we insert the SRL argument label in the corresponding slot
in the joint label.
For predicates sharing more than one argument,
we observe that most cases are
due to the two predicates sharing all their ARGM relations,
so we augment the (C) label with a binary indicator of
whether or not to propagate all ARGM arguments.
When the two predicates share more than one core argument,
which occurs for around $2\%$ of the argument-sharing predicates,
we randomly select and record one of the shared arguments
in $r_i^{\text{(C)}}$.
A more systematic assignment in such cases
in future work
may lead to further improvement.

As for the backward conversion (joint $\rightarrow$ separate),
the syntactic dependencies can be directly decoupled from the joint label,
and we build the SRL relations in three steps:
we first identity all the (D) and (R) dependency relations;
then, with a top-down traversal of the tree,
we identify the shared argument relations through (C) labels;
finally, we rebuild the span boundaries using a rule-based approach.
Top-down traversal is necessary to allow further propagation of arguments.
It allows us to cover some of the less common cases
through multiple argument sharings,
e.g., the fourth example in \reftab{tab:patterns}.
When a (C) label $(a,b)$ is invalid\footnote{
This should not happen in the oracle conversion
but may occur in model predictions.
}
in that the syntactic governor does not have an argument with label $a$,
we simply ignore this (C) label.
In reconstructing the span boundaries,
we distinguish among different types of arguments.
For (D)-type arguments,
we directly take the entire subtrees dominated by
the head words of the arguments.
For (R)-type arguments,
we adopt language-specific heuristics:\footnote{
The simple subtree approach does not apply to
reconstructing (R)-type arguments since, by definition,
the subtree of an (R)-type argument will contain its predicate,
which contradicts data annotations.
Our heuristics are designed to support a span-based evaluation,
and span reconstruction can be omitted
if one focuses on a dependency-based evaluation.
}
in English, when the argument (syntactic head)
is to the left of the predicate (syntactic child),
as commonly happens in relative clause structures,
we include all of the argument's children subtrees
to the left of the predicate;
when the argument is to the right,
which usually happens when the predicate is in participle form,
we define the right subtree of the argument as its span.
For (C)-type arguments,
we reuse the span boundaries of the shared arguments.

\begin{table}[t]
\centering
\small
\begin{tabular}{l|ccc}
\toprule
&
P
&
R
&
F
\\
\midrule
English
&
$99.7$
&
$98.3$
&
$99.0$
\\
Chinese
&
$97.8$
&
$96.8$
&
$97.3$
\\
\bottomrule
\end{tabular}
\caption{
Oracle back-and-forth conversion results
on the training splits.
}
\label{tab:oracle}
\end{table}

\reftab{tab:oracle} shows the oracle results
of our back-and-forth conversion strategies on the training data.
We take gold-standard syntactic and SRL annotations
and convert them into joint-label representations.
Then, we reconstruct the SRL relations
through our backward conversion and
measure span-based exact match metrics.
Our procedures can faithfully reconstruct
most of the SRL relations for both
English and Chinese data.\footnote{
The English oracle F1 score is higher than
the combined (D)+(C)+(R) occurrences of $98\%$.
This is because
(1) our method is precision-focused to minimize error propagation in prediction;
recall loss of $1.7\%$ is a direct reflection of
the unaccounted less-frequent structures,
and (2) many arguments, e.g.,
the fourth most frequent case in \reftab{tab:patterns},
can be reconstructed through the propagation of (C)-type labels.
}
English sees a higher oracle score than Chinese.
We attribute this result to the synchronization effort
between the syntactic and SRL annotations
during the evolution of English PropBank \citep{babko-malaya+06,bonial+17}.

\subsection{Models}

Given that SRL can be reduced to a dependency parsing task
with an extended label space, our model replicates and adapts
that of a dependency parser.
We follow the basic design of \citet{dozat-manning17},
but instead of using LSTMs as input feature extractors,
we opt for Transformer encoders \citep{vaswani+17},
which have previously been shown to be successful in
constituency parsing \citep{kitaev-klein18,kitaev+19},
dependency parsing \citep{kondratyuk-straka19},
and SRL \citep{tan+18a,strubell+18}.
Next, we score all potential attachment pairs
and dependency and SRL relations with the token-level representations
through deep biaffine transformation \citep{dozat-manning17}.
After the dependency parsing decoding process,
we retrieve the syntactic parse trees and SRL structures
via our backward conversion algorithm.

Formally, we associate each token position
with a context-sensitive representation by
\begin{equation*}
\left[\vecs{x_0},\vecs{x_1},\ldots,\vecs{x_n}\right]=\text{Transformer}\left(w_0,w_1,\ldots,w_n\right),
\end{equation*}
where $w_0$ denotes the root symbol for the dependency parse tree,
and the inputs to the Transformer network are pretrained GloVe embeddings \citep{pennington+14}.
Alternatively, we can fine-tune a pre-trained
contextualized
feature extractor such as BERT \citep{devlin+19}:\footnote{
In this case,
we use the final-layer vector of the last sub-word unit for each word
as its representation
and the vector from the prepended \texttt{[CLS]} token
for the root symbol.
}
\begin{equation*}
\left[\vecs{x_0},\vecs{x_1},\ldots,\vecs{x_n}\right]=\text{BERT}\left(\texttt{[CLS]},w_1,\ldots,w_n\right).
\end{equation*}

Next, the same representations $\vecs{x}$ serve as
inputs to five different scoring modules,
one for dependency attachment,
one for syntactic labeling,
and three modules for the newly-introduced SRL-related labels.
All of the scoring modules use a deep biaffine (DBA) scoring function
introduced by \citet{dozat-manning17}
that is widely used in syntactic parsing \citep{dozat+17,shi+17,shi-lee18},
semantic dependency parsing \citep{dozat-manning18}
and SRL \citep{strubell+18}.
For an ordered pair of input vectors $\vecs{x_i}$ and $\vecs{x_j}$,
an $r$-dimensional DBA transforms each vector
into a $d$-dimensional vector
with multi-layer perceptrons and
then outputs an $r$-dimensional vector
$\vecs{z}_{ij}=\text{DBA}(\vecs{x_i},\vecs{x_j})$, where
\begin{equation*}
\vecs{z}_{ijk}=\left[\text{MLP}^{I}(\vecs{x_i});1\right]^{\top} \vecs{U}_k \left[\text{MLP}^{J}(\vecs{x_j});1\right],
\end{equation*}
$\vecs{U}\in \mathbb{R}^{r\times (d+1)\times (d+1)}$,
$[;1]$ appends an element of $1$ to the end of the vector,
and $\text{MLP}^I$ and $\text{MLP}^J$ are
two separate multi-layer perceptrons
with non-linear activation functions.
Following \citet{dozat-manning17},
we model dependency attachment probabilities
with a $1$-dimensional DBA function:
\begin{equation*}
P(h_j=i)\propto \text{exp}(\text{DBA}^\textsc{Att}(\vecs{x_i},\vecs{x_j})).
\end{equation*}
For syntactic labels from vocabulary $V^{\textsc{Syn}}$,
we use a $|V^{\textsc{Syn}}|$-dimensional DBA function:
\begin{equation*}
P(r_j^\textsc{Syn}=V^{\textsc{Syn}}_t)\propto \text{exp}(\text{DBA}^\textsc{Syn}_t(\vecs{x_{h_j}},\vecs{x_j})).
\end{equation*}
The three semantic label components
$r^{\text{(D)}}$, $r^{\text{(C)}}$, and $r^{\text{(R)}}$
are modeled similarly to $r^{\textsc{Syn}}$.

All the above components
are separately parameterized but
they share the same feature extractor (Transformer or BERT).
We train them with locally-normalized log-likelihood as objectives.
During inference,
we use a projective\footnote{
The choice of a projective decoder is motivated by
the empirical fact that both English and Chinese dependency trees
are highly projective.
One may consider a non-projective decoder when adapting to other languages.
}
maximum spanning tree algorithm \citep{eisner96,eisner-satta99}
for unlabeled dependency parsing
and then select the highest-scoring component label
for each predicted attachment and each component.\footnote{
Structured and global inference that considers the interactions
among all relation labels
is a promising future direction.
}

\section{Experiments}

We evaluate on two datasets from
OntoNotes 5.0 \citep{hovy+06} on English and Chinese.
Similar to \refsec{sec:pilot},
we adopt the CoNLL 2012 dataset splits.
To isolate the effects of predicate identification
and following most existing work on SRL,
we provide our models with pre-identified predicates.
We report median performance across $5$ runs
of different random initialization
for our models and our replicated reference models.
Implementation details are provided in Appendix \refsec{app:implementation}.

\paragraph{Main Results}

\begin{table}[t]
\centering
\begin{small}
\begin{tabular}{lccc}
\toprule
\multicolumn{1}{c}{English}
&
P
&
R
&
F1
\\
\midrule
\citet{fitzgerald+15} & $80.9$ & $78.4$ & $79.6$ \\
\citet{he+17} & $81.7$ & $81.6$ & $81.7$ \\
\citet{he+18b} & -- & -- & $82.1$ \\
\citet{tan+18a} & $81.9$ & $83.6$ & $82.7$ \\
\citet{ouchi+18} & $84.4$ & $81.7$ & $83.0$ \\
\citet{swayamdipta+18a} & $85.1$ & $82.6$ & $83.8$ \\
BIO-CRF baseline & $83.4$ & $83.6$ & $83.5$ \\
Ours & $83.3$ & $83.0$ & $83.2$ \\
\midrule
\multicolumn{4}{l}{\emph{with pre-trained contextualized feature extractors}}
\\
\citet{peters+18} & -- & -- & $84.6$ \\
\citet{he+18b} & -- & -- & $85.5$ \\
\citet{ouchi+18} & $87.1$ & $85.3$ & $86.2$ \\
\citet{li+19} & $85.7$ & $86.3$ & $86.0$ \\
\citet{li+20a} & $86.4$ & $86.8$ & $86.6$ \\
BIO-CRF baseline & $86.4$ & $87.1$ & $86.7$ \\
Ours & $85.9$ & $85.6$ & $85.8$ \\
\bottomrule
\end{tabular}

\vspace{6pt}

\begin{tabular}{lccc}
\toprule
\multicolumn{1}{c}{Chinese}
&
P
&
R
&
F1
\\
\midrule
BIO-CRF baseline & $74.3$ & $71.1$ & $72.7$ \\
Ours & $71.7$ & $71.4$ & $71.6$ \\
\midrule
\multicolumn{4}{l}{\emph{with pre-trained contextualized feature extractors}}
\\
BIO-CRF baseline & $80.2$ & $81.1$ & $80.6$ \\
Ours & $79.6$ & $79.3$ & $79.5$ \\
\bottomrule
\end{tabular}

\end{small}

\caption{
Non-ensemble CoNLL 2012 test set results on both the English and the Chinese datasets.
}
\label{tab:results}
\end{table}

\reftab{tab:results} reports the evaluation results.
We compare our method with multiple state-of-the-art methods,
including BIO-tagging \citep{tan+18a,peters+18},
span-based \citep{ouchi+18,li+19},
semi-Markov CRF \citep{swayamdipta+18a}
and structured tuning \citep{li+20a}.
We also implement a strong BIO-tagging model
trained with a CRF loss as our baseline model (BIO-CRF),
which has identical feature extractors as our proposed method.\footnote{
See, for example, \citet{he+17} for a BIO-tagging formulation of SRL.
}
Results show that our models are competitive
with the state-of-the-art models,
even though our method reduces SRL to syntactic dependency parsing.
Our models slightly underperform the BIO-CRF baseline models on English,
and the gap is larger on Chinese.\footnote{
An anonymous reviewer hypothesizes that
the accuracy loss may also be explained by the label sparsity
introduced by our joint label scheme.
}
This can be attributed to the higher back-and-forth conversion loss on the Chinese data.
We observe no significant difference in dependency parsing accuracy
when training the \citeauthor{dozat-manning17}'s (\citeyear{dozat-manning17}) parser alone
versus jointly training with our SRL labels.

Additionally, our models make predictions
for all predicates in a given sentence at the same time
through $O(n)$ joint syntacto-semantic labels with identical features,
while most other competitive methods either use
different features extracted for different predicates \citep{tan+18a,ouchi+18,swayamdipta+18a},
effectively requiring executing feature extraction multiple times,
or require scoring for all $O(n^2)$ or $O(n^3)$ possible predicate-argument pairs\footnote{
$O(n^3)$ results from considering all combinations of
predicates, span start points and end points.
In practice, however, \citet{li+19} apply pruning to reduce number of candidates.
} \citep{strubell+18,li+19}.
In our experiments, our models are $40\%$ faster than the BIO-CRF baseline on average.

\paragraph{Results Broken Down by Argument Type}

\begin{table}[t]
\centering
\small
\begin{tabular}{lrccc}
\toprule
\multicolumn{1}{c}{Label}
&
\multicolumn{1}{c}{Count}
&
BIO-CRF
&
Ours
&
+BERT
\\
\midrule
ARG0 & $11{,}444$ & $90.8$ & $90.4$ & $91.8$ \\
ARG1 & $18{,}216$ & $86.0$ & $85.7$ & $88.7$ \\
ARG2 &  $6{,}429$ & $80.1$ & $78.4$ & $83.7$ \\
\midrule
ARGM-TMP & $3{,}724$ & $83.4$ & $83.8$ & $86.6$ \\
ARGM-ADV & $2{,}089$ & $65.0$ & $63.9$ & $66.5$ \\
ARGM-DIS & $2{,}378$ & $82.1$ & $82.4$ & $83.1$ \\
ARGM-MOD & $1{,}844$ & $97.8$ & $98.0$ & $97.8$ \\
\midrule
Overall  & $53{,}906$ & $83.5$ & $83.0$ & $85.9$ \\
\bottomrule
\end{tabular}
\caption{
Per-type performance on the English dev set.
}
\label{tab:perlab}
\end{table}

\reftab{tab:perlab} presents per-label F1 scores
comparing our baseline model with our proposed method.
Our method exhibits a similar overall performance
to the baseline BIO-CRF model.
Most of the difference
is materialized on
ARG2 and ARGM-ADV.
Previous work in the literature finds that
these labels are highly predicate-specific
and known to be hard to predict \cite{he+17}.
We further observe that pretrained feature extractors (BERT)
tend to improve the most with respect to these two labels.

\paragraph{Effect of Different Components}

\begin{table}[t]
\centering
\small
\begin{tabular}{l|cc}
\toprule
\multirow{2}{*}{Setting}
&
\multicolumn{2}{c}{F1 Score}
\\
&
GloVe
&
BERT
\\
\midrule
All predicted
& $83.0$ & $85.9$ \\
+ Gold syntax
& $88.9$ & $90.0$ \\
\quad + Gold (D)
& $97.2$ & $97.3$ \\
\quad + Gold (R) (C)
& $90.1$ & $91.1$ \\
\midrule
Upperbound (Oracle)
&
\multicolumn{2}{c}{$98.9$} \\
\bottomrule
\end{tabular}
\caption{
F1 performance on the English dev set if parts of the components are replaced by oracle, as an indicator of potential further gains from each component.
}
\label{tab:ablation}
\end{table}

\reftab{tab:ablation} summarizes the results
when one or more components of our models
are replaced by gold-standard labels.
As expected, it is crucial to predict the syntactic trees correctly:
failure to do so amounts to $35\%$ or $29\%$ of errors
with or without pretrained feature extractors.
Accuracy of (D)-type SRL relations has an even larger impact
on the overall performance:
it is responsible for half of the errors.
This indicates that argument labeling is
a harder sub-task than syntactic parsing.
Further, we observe that the benefits of
pretrained feature extractors mostly stem from
improved accuracies of the syntactic component.
Even with pretrained BERT features,
semantic components remain challenging.

\section{Related Work}
\label{sec:related}

\paragraph{SRL and syntax}
From the time the SRL task was first introduced
\citep{gildea-jurafsky02,gildea-palmer02,marquez+08,palmer+10},
syntax has been shown to be a critical factor in system performance.
Most models use syntactically-derived features
\citep[\emph{inter alia}]{pradhan+05,punyakanok+05,swanson-gordon06,johansson-nugues08,toutanova+08,xue08,zhao+09}
and syntax-based candidate pruning \citep{punyakanok+08}.
There have been many approaches for joint syntactic parsing and SRL models,
including approximate search \citep{johansson09}
and dual decomposition \citep{lluis+13}
to resolve feature dependencies,
and synchronous parsing to simultaneously
derive the (disjoint) syntactic and SRL structures
\citep{henderson+08,li+10,henderson+13,swayamdipta+16}.
In contrast, our work unifies the two representations
into common structures.

\paragraph{Joint labels}
The idea of using joint labels for performing
both syntactic and semantic tasks is similar to
that of function parsing \citep{merlo-musillo05,gabbard+06,musillo-merlo06}.
\citet{ge-mooney05} use joint labels for semantic parsing as well.
Earlier approaches for SRL have considered
joint syntactic and semantic labels.
Due to lack of characterization of the common structures,
most work either focuses on the subtask of argument identification \citep{yi-palmer05},
predicts the set of all SRL labels for each argument
and links them to predicates in a second stage \citep{merlo-musillo08},
or models joint labels independently for each predicate \citep{samuelsson+08,lluis-marquez08,morante+09,rekabysalama-menzel19}.
Instead, our work aims at extracting
all predicate-argument structures from a sentence.
Our joint label design is related to that of \citet{qiu+16}.
They annotated a Chinese SRL corpus from scratch with
a similar label scheme, while in this paper,
we show that it is possible to extract such
joint labels from existing data annotations.

\paragraph{Tree approximation}
In the task of semantic dependency parsing \citep{oepen+14},
dependency structures are used to model
more aspects of semantic phenomena
than predicate-argument structures,
and the representations are more general directed acyclic graphs.
These graphs can be approximated by trees
\citep{du+14,schluter+14,schluter15} such that
tree-based parsing algorithms become applicable.
Unlike this line of research,
we limit ourselves to the \emph{given} syntactic trees,
as opposed to finding the \emph{optimal approximating} trees,
and we focus on the close relations between syntax and SRL.

\paragraph{Dependency-based SRL}
Although predicate-argument structures are traditionally defined
in constituency terms,
dependency-based predicate-argument analysis \citep{hacioglu04,fundel+07}
has been popularized through the CoNLL 2008 and 2009 shared tasks \citep{surdeanu+08,hajic+09}
and has been adopted by recent proposals of decompositional semantics \citep{white+17}.
\citet{choi-palmer10} consider reconstructing
constituency-based representations from dependency-based analysis.
We confirm their findings that through a few heuristics,
the reconstruction can be done faithfully.

\paragraph{Neural SRL}
The application of neural models to SRL
motivates the question of whether modeling syntax
is still necessary for the task \citep{he+17}.
Similar to non-neural models, syntactic trees
are used to construct features \citep{roth-lapata16a,kasai+19,wang+19,xia+19,zhang+19}
and to prune candidates \citep{tackstrom+15,he+18a,he+19}.
Alternatively, they are used to determine network structures \citep{marcheggiani-titov17,li+18},
including tree-LSTM, graph convolutional networks \citep{niepert+16}
and syntax-aware LSTM \citep{qian+17}.
On the other hand, syntax-agnostic models \citep{collobert-weston07,zhou-xu15,cai+18,he+18b,tan+18a,li+19}
have shown competitive results.
Our results contribute to the ongoing debate
by adding further evidence that
the two tasks are deeply-coupled.
Future work may further explore how much syntactic knowledge
has been implicitly obtained in the
apparently syntax-agnostic models.

\paragraph{Multi-task learning}
Our models share neural representations
across the syntactic and the SRL labelers.
This is an instance of multi-task learning \citep[MTL;][]{caruana93,caruana97}.
MTL has been successfully applied to SRL \citep{collobert-weston08,collobert+11,shi+16}
in many state-of-the-art systems \citep{strubell+18,swayamdipta+18a,cai-lapata19,xia+19a}.
A potential future extension is to
learn multiple syntactic \citep{sogaard-goldberg16}
and semantic representations \citep{peng+17a,hershcovich+18}
beyond dependency trees and PropBank-style SRL
at the same time.

\section{Conclusion}

Linguistic theories assume a close relationship
between the realization of semantic arguments and syntactic configurations.
This work provides a detailed analysis
of the syntactic structures of PropBank-style SRL
and reveals that three common syntactic patterns
account for $98\%$ of annotated SRL relations
for both English and Chinese data.
Accordingly, we propose to reduce the task of SRL
to syntactic dependency parsing
through back-and-forth conversion to and from a joint label space.
Experiments show that
dependency parsers achieve competitive results on PropBank-style SRL
with the state of the art.

This work shows promise of a syntactic treatment of SRL
and opens up possibilities of applying
existing dependency parsing techniques to SRL.
We invite future research into further integration of
syntactic methods into shallow semantic analysis in other languages
and other formulations, such as frame-semantic parsing,
and other semantically-oriented tasks.

\section*{Acknowledgements}

We thank the anonymous reviewers for their insightful reviews,
and
Sameer Bansal,
Sam Brody,
Prabhanjan Kambadur,
Lillian Lee,
Daniel Preo{\c t}iuc-Pietro,
Mats Rooth,
Amanda Stent,
and Chen-Tse Tsai for
discussion
and comments.
Tianze Shi acknowledges support from Bloomberg's Data Science Ph.D. Fellowship.

\bibliography{ref}
\bibliographystyle{acl_natbib}

\clearpage
\appendix

\section{Implementation Details}
\label{app:implementation}

We use two types of encoders in our models: randomly-initialized Transformer \citep{vaswani+17} networks and pre-trained BERT \citep{devlin+19}.
For Transformer networks, the input representations are concatenations of $100$-dimensional randomly-initialized word embeddings, $100$-dimensional pre-trained GloVe \citep{pennington+14} embeddings, $16$-dimensional predicate indicator embeddings and $128$-dimensional positional embeddings.
The Transformer networks have $8$ self-attention and feed-forward layers.
Each self-attention layer has $8$ attention heads, and each feed-forward layer has a dimensionality of $2048$.
For BERT models, we fine-tune the pretrained \textsc{base} model by \citet{devlin+19} and \citet{wolf+19}.

The decoders consist of an unlabeled attachment scorer and several labeling components for the syntactic dependencies and SRL relations.
The design and hyperparameters follow that of \citet{dozat-manning17}.
The biaffine scoring function for the unlabeled attachment scorer has a dimensionality of $400$, while each labeling component has $100$ dimensions.
For our baseline, we build on top of \citeauthor{tan+18a}'s (\citeyear{tan+18a}) BIO tagging model and further add a CRF-based decoding layer following \citet{yang+18}.
Contextualized representation at each token's position is passed through a multi-layer perceptron with one hidden layer consisting of $256$ hidden units and PReLU \citep{he+15} activation function to obtain the scores for each tag.

$64$ training instances ($16$ when using BERT) are grouped into a minibatch, and the gradients are clipped \citep{pascanu+13} at $5.0$.
We use Adam \citep{kingma-ba15} optimizer with $\beta_1=0.9$, $\beta_2=0.999$ and $\epsilon=1\times 10^{-8}$.
When using GloVe embeddings and Transformers, we set the learning rate to be $1\times 10^{-4}$;
when fine-tuning BERT, the learning rate is lowered to $1\times 10^{-5}$.
Learning rates are multiplied by $0.1$ once the development performance stops increasing for $5$ epochs.
All the models are trained until the learning rates are lowered three times and the performance plateaus on the development sets.
Our implementation is based on PyTorch \citep{paszke+17}.

On a single V100 GPU, the baseline BIO-CRF model parses $96.4$ sentences/sec and our proposed model processes at $159.1$ sentences/sec on average.

Throughout our experiments,
all the hyperparameters are taken directly
from relevant suggestions in previous literature \citep{dozat-manning17,tan+18a,kitaev+19}
without tuning.
An extensive hyperparameter search may lead to further accuracy improvements.

\section{Additional Model Analysis}
\label{app:analysis}

\subsection{Training without Gold Syntactic Trees}

Our method leverages the gold-standard dependency trees in the \emph{training} data to design high-fidelity back-and-forth conversion algorithms.
\reftab{tab:jackknife} considers a scenario where we do not have access to such gold trees during \emph{training}:
we jackknife the data into $8$ folds, train parsers using $7$ folds and predict trees on the remaining fold.
Our models show similar F1 scores under this condition as that of using gold trees,
while the recall is traded for precision since our conversion method is precision-focused.

This is not a realistic scenario given that
existing PropBank-style SRL annotations are
all based on syntax,
so as a matter of practice
we always have access to gold trees during \emph{training}.
Nonetheless, these experiments
point to the viability of using predicted trees
in practice without incurring a significant loss in F1 scores.

\begin{table}[t]
\centering
\small
\begin{tabular}{llccc}
\toprule
                        & Trained with         & P & R & F \\
\midrule
\multirow{2}{*}{Oracle}
                        & Gold      & $99.7$ & $98.2$ & $98.9$ \\
                        & Predicted & $99.6$ & $93.2$ & $96.3$ \\
\midrule
\multirow{2}{*}{GloVe}
                        & Gold      & $83.2$ & $82.9$ & $83.0$ \\
                        & Predicted & $84.7$ & $80.6$ & $82.6$ \\
\midrule
\multirow{2}{*}{BERT}
                        & Gold      & $86.2$ & $85.5$ & $85.8$ \\
                        & Predicted & $86.9$ & $83.2$ & $85.1$ \\
\bottomrule
\end{tabular}
\caption{
English back-and-forth oracle and dev set results using \emph{gold-standard} dependency trees versus \emph{predicted} trees as \emph{training} data.
}
\label{tab:jackknife}
\end{table}

\subsection{Accuracies by SRL Relation Types}

In \reftab{tab:pertype}, we break down the accuracies by the syntactic patterns of the SRL relations.
Compared with our baseline, a replication of \citet{tan+18a},
our models achieves higher or competitive results on (D)-type and (R)-type SRL relations.
These two types establish a direct or reverse semantic relation with respect to the syntactic structure.
In contrast, the (C)-type relations require accurate predictions of sibling relations as well as at least two SRL-related labels and are thus more prone to error propagation.
We hypothesize that global scoring of the dependency structures can alleviate this issue, and we leave that to future work.

\begin{table}[t]
\centering
\begin{tabular}{c|cc|cc}
\toprule
\multirow{2}{*}{Type}
&
\multicolumn{2}{c|}{English}
&
\multicolumn{2}{c}{Chinese}
\\
&
Baseline
&
Ours
&
Baseline
&
Ours
\\
\midrule
(D) & $88.1$ & $88.1$ & $82.8$ & $83.4$ \\
(C) & $80.3$ & $76.6$ & $53.3$ & $50.0$ \\
(R) & $78.7$ & $79.3$ & $46.3$ & $46.8$ \\
\midrule
Overall  & $83.5$ & $83.0$ & $74.1$ & $72.9$ \\
\bottomrule
\end{tabular}
\caption{
Per-pattern F1 scores on the dev sets.
}
\label{tab:pertype}
\end{table}

\subsection{Learning Curve}

In \reftab{tab:learning-curve}, we train the models with varying amounts of training data.
With GloVe embeddings, our models exhibit higher performance when training data is limited,
as compared with the corresponding baselines.
When the pre-trained BERT feature extractor is used,
both the baseline and our model require far less data to reach similar levels of performance.
Our model shows significant improvement when the amount of training data is extremely limited ($1\%$),
and the baseline edges out for the other two settings ($3\%$ and $10\%$).

\begin{table}[t]
\centering
\begin{tabular}{c|c@{\hspace{0.5em}}c@{\hspace{0.5em}}c|c@{\hspace{0.5em}}c@{\hspace{0.5em}}c}
\toprule
\multirow{2}{*}{}
&
\multicolumn{3}{c|}{GloVe}
&
\multicolumn{3}{c}{BERT}
\\
&
Base.
&
Ours
&
$\Delta$
&
Base.
&
Ours
&
$\Delta$
\\
\midrule
$10\%$ & $71.3$ & $72.4$ & $+1.1$ & $82.3$ & $81.5$ & $-0.8$ \\
$3\%$ & $59.9$ & $62.0$ & $+2.1$ & $78.2$ & $77.6$ & $-0.6$ \\
$1\%$ & $47.5$ & $48.7$ & $+1.2$ & $69.3$ & $73.2$ & $+3.9$ \\
\bottomrule
\end{tabular}
\caption{
English dev F1 scores, when trained with different percentages of the training data.
}
\label{tab:learning-curve}
\end{table}

\section{Additional English Data Analysis}
\label{app:eng}

Among the three common patterns, (D)-type SRL relations are the most frequent and easiest to understand.
In this section, we provide additional examples to shed light on (C)-type and (R)-type relations.
We also show some sentences with more complex syntactic phenomena than what can be handled by our joint-label scheme.
In all the examples, we boldface the predicates, underline the head words of the arguments,
and highlight only the shortest dependency paths connecting them.

\subsection{(C)-Type Relations}

The (C)-type relations are most frequently used in ARG0 ($55\%$) and ARG1 ($19\%$) relations, in contrast to (D)-type relations, where the percentages are much lower ($34\%$ and $17\%$ respectively).
This can be explained by the fact that a lot of (C)-type relations are used in control and raising verb constructions.
A second major construction associated with (C)-type relations is conjunction,
which shares either core or peripheral arguments among the conjuncts.
The most common dependency relation labels connecting the common parents and the predicates are:
``xcomp'' ($39\%$),
``conj'' ($37\%$),
``vmod'' ($9\%$),
and ``dep'' ($6\%$).

``xcomp'' signifies control/raising structures.
Popular common parent words (the control/raising verbs) include ``want'', ``expect'', ``continue'', ``begin'', etc.

``conj'' represents a coordination structure.
Since the first conjunct is a syntactic head of other conjuncts in Stanford Dependencies, any shared argument will result in a (C)-type relation.

``vmod'' denotes non-finite verbal modifiers whose missing subjects can often be found in the main clauses.
For example:
\begin{exe}
\ex
\underline{We} use all wisdom to \textbf{counsel} every person.
\vspace{5pt}
\\
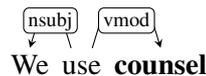
\begin{dependency}
\begin{deptext}
\underline{We} \& use \& \textbf{counsel}\\
\end{deptext}
\depedge[edge height=2ex]{2}{1}{nsubj}
\depedge[edge height=2ex]{2}{3}{vmod}
\end{dependency}
\end{exe}

A lot of problematic instances of ``dep'' can be attributed to failures of constituency-to-dependency conversion, where it should have been recognized as a relation corresponding to another construction.
For example:
\begin{exe}
\ex
\underline{He} calls \ldots and \textbf{pops} in every once in a while.
\vspace{5pt}
\\
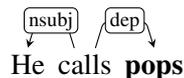
\begin{dependency}
\begin{deptext}
\underline{He} \& calls \& \textbf{pops}\\
\end{deptext}
\depedge[edge height=2ex]{2}{1}{nsubj}
\depedge[edge height=2ex]{2}{3}{dep}
\end{dependency}
\end{exe}

\subsection{(R)-Type Relations}

(R)-type relations are frequently used in relative clauses, as ``rcmod'' accounts for $47
\%$ of the syntactic relations connecting the predicates and the arguments.
For examples:
\begin{exe}
\ex
\ldots another \underline{group} that is always \textbf{trying} to \ldots
\vspace{5pt}
\\
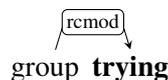
\begin{dependency}
\begin{deptext}
\underline{group} \& \textbf{trying}\\
\end{deptext}
\depedge[edge height=2ex]{1}{2}{rcmod}
\end{dependency}
\end{exe}
\begin{exe}
\ex
\ldots outer \underline{part} of the nursery where we were \textbf{waiting} \ldots
\vspace{5pt}
\\
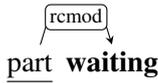
\begin{dependency}
\begin{deptext}
\underline{part} \& \textbf{waiting}\\
\end{deptext}
\depedge[edge height=2ex]{1}{2}{rcmod}
\end{dependency}
\end{exe}

The second most common construction involves ``vmod'' ($28\%$). Different from the ``vmod'' relations involved in (C)-type relations, the non-finite clauses usually modify noun phrases in (R)-type relations. For examples:
\begin{exe}
\ex
\ldots developed \ldots management \underline{consultants} to \textbf{go} out \ldots
\vspace{5pt}
\\
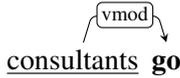
\begin{dependency}
\begin{deptext}
\underline{consultants} \& \textbf{go}\\
\end{deptext}
\depedge[edge height=2ex]{1}{2}{vmod}
\end{dependency}
\end{exe}
\begin{exe}
\ex
The \underline{administration}, \textbf{hoping} to de-escalate the violence, is appealing to both sides.
\vspace{5pt}
\\
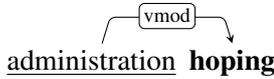
\begin{dependency}
\begin{deptext}
\underline{administration} \& \textbf{hoping}\\
\end{deptext}
\depedge[edge height=2ex]{1}{2}{vmod}
\end{dependency}
\end{exe}

The third most common type of cases involves participial adjectives, using ``amod'' syntactic relation ($17\%$).
Since the verb is modifying the noun as an adjective, the syntactic dependency and the semantic relation are reversed. For example:
\begin{exe}
\ex
\ldots a fact finding American \textbf{led} \underline{committee} \ldots
\vspace{5pt}
\\
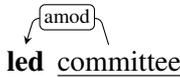
\begin{dependency}
\begin{deptext}
\textbf{led} \& \underline{committee}\\
\end{deptext}
\depedge[edge height=2ex]{2}{1}{amod}
\end{dependency}
\end{exe}

\subsection{Others}

The other constructions besides the three most common patterns are a mixture of data annotation errors, constituency-to-dependency failures, and combinations of the frequent patterns.

If the argument is shared with other predicates along the dependency path, then our conversion algorithm can recover the SRL relation through multiple (C)-type labels.
For example, in the following sentence, the argument ``I'' is shared across three predicates ``trying'', ``help'' and ``fix'' as ARG0's.
\begin{exe}
\ex
\underline{I}'ve been \textbf{trying} to \textbf{help} him \textbf{fix} \ldots
\vspace{5pt}
\\
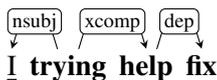
\begin{dependency}
\begin{deptext}
\underline{I} \& \textbf{trying} \& \textbf{help} \& \textbf{fix}\\
\end{deptext}
\depedge[edge height=2ex]{2}{1}{nsubj}
\depedge[edge height=2ex]{2}{3}{xcomp}
\depedge[edge height=2ex]{3}{4}{dep}
\end{dependency}
\end{exe}

Annotation inconsistencies can result in rare patterns beyond the scope of the current design of our joint label. For example, in the following sentence, the SRL annotation decides that ``the Museum of Modern Art'' is ARGM-LOC of ``listed'', making the predicate a grandparent of the argument.
A simple fix that simply includes the preposition ``in'' as part of the argument span (as is annotated in most other examples) will change this case into a (D)-type relation.
\begin{exe}
\ex
Now your name is \textbf{listed} in the \underline{Museum} of Modern Art.
\vspace{5pt}
\\
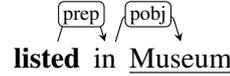
\begin{dependency}
\begin{deptext}
\textbf{listed} \& in \& \underline{Museum}\\
\end{deptext}
\depedge[edge height=2ex]{1}{2}{prep}
\depedge[edge height=2ex]{2}{3}{pobj}
\end{dependency}
\end{exe}

\section{Chinese Data Analysis}
\label{app:chinese}

Despite the fact that Chinese and English are very different languages from two distinctive language families, they exhibit similar distributions of patterns when it comes to the syntactic patterns of SRL relations.
The three most common types, (D)-, (C)- and (R)-type relations, account for over $98\%$ of all annotated predicate-argument relations.
In the following examples, \textsc{ba} denotes a \emph{ba} construction, \textsc{de} refers to a \emph{de} particle, and \textsc{classifier} represents Chinese measure words for quantity expressions \citep{huang+09}.

\subsection{(D)-Type Relations}
Similarly to English, most Chinese SRL relations parallel the syntactic counterparts.
For example, in the following sentence, each of the three arguments of the predicate corresponds to a direct dependent in the dependency structure.
\begin{exe}
\ex
\gll
\underline{\chn{浙江}} \chn{把} \chn{特色} \underline{\chn{产业区}} \textbf{\chn{作为}} \chn{经济} \chn{发展} \chn{的} \chn{战略} \underline{\chn{选择}}\\
\underline{ZheJiang} \textsc{ba} featured \underline{industrial-zones} \textbf{take-as} economic development \textsc{de} strategic \underline{choice}\\
\trans
``Zhejiang uses its featured industrial zones as a strategic choice for economic development.''
\vspace{5pt}
\\
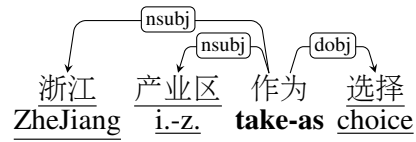
\begin{dependency}
\begin{deptext}
\underline{\chn{浙江}} \& \underline{\chn{产业区}} \& \textbf{\chn{作为}} \& \underline{\chn{选择}}\\
\underline{ZheJiang} \& \underline{i.-z.} \& \textbf{take-as} \& \underline{choice}\\
\end{deptext}
\depedge[edge height=4ex]{3}{1}{nsubj}
\depedge[edge height=2ex]{3}{2}{nsubj}
\depedge[edge height=2ex]{3}{4}{dobj}
\end{dependency}
\end{exe}
``i.-z.'' is short for ``industrial-zones''.

\subsection{(C)-Type Relations}

In a (C)-type relation,
the most frequent syntactic labels between the common parent and the predicate are ``dep'' ($37.5\%$), ``conj'' ($34.4\%$) and ``mmod'' ($9.6\%$).
``conj'' denotes coordinations, as discussed in the English data analysis section.
Unlike English data, the Chinese annotations use a large percentage of ``dep'' relations.
A closer inspection reveals that most of the instances correspond to open clausal complements (``xcomp'') and coordinations (``conj''). (See English data analysis section for analysis.)

\begin{exe}
\ex
\gll
\underline{\chn{我}} \chn{真的} \chn{非常} \chn{努力} \chn{地} \chn{工作} \ldots \chn{以} \textbf{\chn{减少}} \ldots\\
\underline{I} really very diligently \textsc{de} work \ldots to \textbf{reduce} \ldots\\
\trans
``I work very diligently to reduce \ldots''
\vspace{5pt}
\\
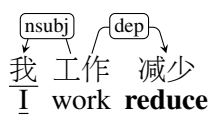
\begin{dependency}
\begin{deptext}
\underline{\chn{我}} \& \chn{工作} \& \textbf{\chn{减少}}\\
\underline{I} \& work \& \textbf{reduce}\\
\end{deptext}
\depedge[edge height=2ex]{2}{1}{nsubj}
\depedge[edge height=2ex]{2}{3}{dep}
\end{dependency}
\end{exe}
\begin{exe}
\ex
\gll
\underline{\chn{我}} \chn{能} \chn{接受} \chn{这个} \chn{，} \chn{并且} \chn{能} \textbf{\chn{宣布}} \chn{它}\\
\underline{I} can accept this , and can \textbf{announce} it\\
\trans
``I can accept this and announce it''
\vspace{5pt}
\\
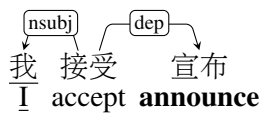
\begin{dependency}
\begin{deptext}
\underline{\chn{我}} \& \chn{接受} \& \textbf{\chn{宣布}}\\
\underline{I} \& accept \& \textbf{announce}\\
\end{deptext}
\depedge[edge height=2ex]{2}{1}{nsubj}
\depedge[edge height=2ex]{2}{3}{dep}
\end{dependency}
\end{exe}

``mmod'' is a dependency relation specific to Chinese.
This label represents modal verb modifiers.
In the Chinese SRL data, many of the modal verbs are annotated as predicates, resulting in (C)-type patterns.
Additionally, ``mmod'' is frequently overloaded with conversions from some multi-verb constructions.
The following sentence shows a common argument of two predicates.
The first one (``should'') is a modal verb, while the second one (``continue'') is often treated as a standalone verb in a multi-verb construction \citep{li-thompson89}.
\begin{exe}
\ex
\gll
\underline{\chn{但是}} \textbf{\chn{要}} \textbf{\chn{继续}} \chn{加大} \chn{改革} \chn{力度}\\
\underline{but} \textbf{should} \textbf{continue} increase reform strength\\
\trans
``but we should continue to strengthen the reforms''
\vspace{5pt}
\\
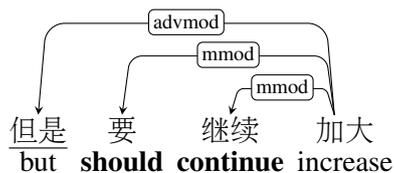
\begin{dependency}
\begin{deptext}
\underline{\chn{但是}} \& \textbf{\chn{要}} \& \textbf{\chn{继续}} \& \chn{加大}\\
\underline{but} \& \textbf{should} \& \textbf{continue} \& increase\\
\end{deptext}
\depedge[edge height=7ex]{4}{1}{advmod}
\depedge[edge height=4.5ex]{4}{2}{mmod}
\depedge[edge height=2ex]{4}{3}{mmod}
\end{dependency}
\end{exe}

\subsection{(R)-Type Relations}
Similar to English, (R)-type relations are frequently used in relative clauses in Chinese as well.
``rcmod'' accounts for $64\%$ of the syntactic relations connecting the predicates and the arguments. For example:
\begin{exe}
\ex
\gll
\chn{一} \chn{栋} \chn{众多} \chn{商户} \textbf{\chn{相连}} \chn{的} \chn{商业} \underline{\chn{楼}}\\
a \textsc{classifier} many merchants \textbf{connect} \textsc{de} commercial \underline{building}\\
\trans
``a commercial building that connects many merchants''
\vspace{5pt}
\\
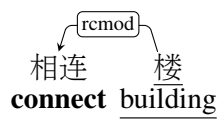
\begin{dependency}
\begin{deptext}
\textbf{\chn{相连}} \& \underline{\chn{楼}}\\
\textbf{connect} \& \underline{building}\\
\end{deptext}
\depedge[edge height=2ex]{2}{1}{rcmod}
\end{dependency}
\end{exe}

Other common constructions include ``mmod'' ($12.9\%$), ``dep'' ($6.6\%$) and ``dobj'' ($5.9\%$).
Complements of modal verbs can result in (R)-type patterns, illustrated as follows:
\begin{exe}
\ex
\gll
\chn{上海} \textbf{\chn{要}} \underline{\chn{建}} \chn{四} \chn{个} \chn{中心}\\
Shanghai \textbf{want} \underline{build} four \textsc{classifier} center\\
\trans
``Shanghai wants to build itself as four centers''
\vspace{5pt}
\\
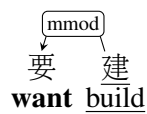
\begin{dependency}
\begin{deptext}
\textbf{\chn{要}} \& \underline{\chn{建}}\\
\textbf{want} \& \underline{build}\\
\end{deptext}
\depedge[edge height=2ex]{2}{1}{mmod}
\end{dependency}
\end{exe}

In (R)-type patterns, ``dobj'' commonly corresponds to light verb constructions where the object nouns are nominalized predicates while the syntactic heads are light verbs without much semantic meaning.
Here we show an example:
\begin{exe}
\ex
\gll
\underline{\chn{进行}} \chn{适当} \textbf{\chn{调整}}\\
\underline{do} adequate \textbf{adjustment}\\
\trans
``adjust adequately''
\vspace{5pt}
\\
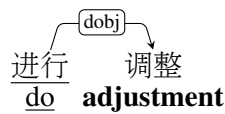
\begin{dependency}
\begin{deptext}
\underline{\chn{进行}} \& \textbf{\chn{调整}}\\
\underline{do} \& \textbf{adjustment}\\
\end{deptext}
\depedge[edge height=2ex]{1}{2}{dobj}
\end{dependency}
\end{exe}

Finally, cases involving ``dep'' relations include miscellaneous data annotation errors and constituency-to-dependency conversion errors.

\subsection{Others}

Similar to English, many of the unaccounted Chinese syntactic patterns of SRL relations are combinations of the three basic patterns.
The following sentence illustrates propagation of an argument through multiple (C)-type structures.
\begin{exe}
\ex
\gll
\underline{\chn{医院}} \textbf{\chn{扩大}} \chn{药品} \chn{和} \chn{医疗} \chn{仪器} \chn{采购} \chn{规模} \chn{从而} \textbf{\chn{压缩}} \chn{单位} \chn{成本} \chn{、} \textbf{\chn{扩大}} \textbf{\chn{服务}}\\
\underline{hospital} \textbf{expand} medicine and medical equipment purchase scale in-order-to \textbf{reduce} unit cost , \textbf{expand} \textbf{service}\\
\trans
``Hospitals expand the scale of purchasing medicines and medical equipments in order to reduce unit costs and to expand their service''
\vspace{5pt}
\\
\begin{dependency}
\begin{deptext}
\underline{\chn{医院}} \& \textbf{\chn{扩大}} \& \textbf{\chn{压缩}} \& \textbf{\chn{扩大}} \& \textbf{\chn{服务}}\\
\underline{hospital} \& \textbf{expand} \& \textbf{reduce} \& \textbf{expand} \& \textbf{service}\\
\end{deptext}
\depedge[edge height=2ex]{2}{1}{nsubj}
\depedge[edge height=2ex]{2}{3}{conj}
\depedge[edge height=2ex]{3}{4}{conj}
\depedge[edge height=2ex]{4}{5}{dobj}
\end{dependency}
\end{exe}

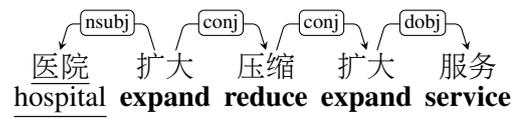
\end{document}